\documentclass{ieeeaccess}
\makeatletter
\def\headerlogo{}
\def\headerlogoall{}
\makeatother
\usepackage{cite}
\usepackage{amsmath,amssymb,amsfonts}
\usepackage{algorithmic}
\usepackage{graphicx}
\usepackage{textcomp}
\usepackage{url}

\usepackage{bm}
\makeatletter
\AtBeginDocument{\DeclareMathVersion{bold}
\SetSymbolFont{operators}{bold}{T1}{times}{b}{n}
\SetSymbolFont{NewLetters}{bold}{T1}{times}{b}{it}
\SetMathAlphabet{\mathrm}{bold}{T1}{times}{b}{n}
\SetMathAlphabet{\mathit}{bold}{T1}{times}{b}{it}
\SetMathAlphabet{\mathbf}{bold}{T1}{times}{b}{n}
\SetMathAlphabet{\mathtt}{bold}{OT1}{pcr}{b}{n}
\SetSymbolFont{symbols}{bold}{OMS}{cmsy}{b}{n}
\renewcommand\boldmath{\@nomath\boldmath\mathversion{bold}}}
\makeatother

\def\BibTeX{{\rm B\kern-.05em{\sc i\kern-.025em b}\kern-.08em
T\kern-.1667em\lower.7ex\hbox{E}\kern-.125emX}}

\usepackage{soul}
\makeatletter
\def\@doi{}
\def\doifont{}
\makeatother
\begin{document}


\title{LGDWT-GS: Local and Global Discrete Wavelet-Regularized 3D Gaussian Splatting for Sparse-View Scene Reconstruction}

\author{
\uppercase{Shima Salehi}\authorrefmark{1},
\uppercase{Atharva Agashe}\authorrefmark{1},
\uppercase{Andrew J. McFarland}\authorrefmark{2},
and \uppercase{Joshua Peeples}\authorrefmark{1}\IEEEmembership{Member, IEEE}
}

\address[1]{Department of Electrical and Computer Engineering, Texas A\&M University, College Station, TX 77843 USA (e-mail: shima.salehi@tamu.edu, atharvagashe22@tamu.edu, jpeeples@tamu.edu)}
\address[2]{Department of Horticultural Sciences and Texas A\&M AgriLife Research Automated Precision Phenotyping Facility, Texas A\&M University, College Station, TX 77843 USA (e-mail: andrew.mcfarland@ag.tamu.edu)}

\tfootnote{This material is based upon work supported by the Texas A\&M University System Nuclear Security Office. Portions of this research were conducted with advanced computing resources provided by Texas A\&M High Performance Research Computing.}



\begin{abstract}
We propose a new method for few-shot 3D reconstruction that integrates global and local frequency regularization to stabilize geometry and preserve fine details under sparse-view conditions, addressing a key limitation of existing 3D Gaussian Splatting (3DGS) models. We also introduce a multispectral greenhouse dataset containing four spectral bands captured from diverse plant species under controlled conditions. Alongside the dataset, we release an open-source benchmarking package that defines standardized few-shot reconstruction protocols for evaluating 3DGS-based methods. Experiments on our multispectral dataset, as well as standard benchmarks, demonstrate that the proposed method achieves sharper, more stable, and spectrally consistent reconstructions than existing baselines. The dataset and code for this work are publicly available at \url{https://github.com/Advanced-Vision-and-Learning-Lab/sparse-view-3dgs-pack}.
\end{abstract}


\titlepgskip=-21pt

\maketitle

\section{Introduction}

Novel view synthesis aims to render unseen viewpoints of a scene from a limited number of images. Neural Radiance Fields (NeRF)~\cite{mildenhall2021nerf} achieve photorealistic quality but rely on dense multi-view supervision and long training times, which limits their use in practical settings such as robotics~\cite{irshad2024neural} and agriculture~\cite{hu2024high} domains where capturing many views is often infeasible. 3DGS~\cite{kerbl20233d} addresses these computational bottlenecks, enabling real-time rendering and efficient optimization. However, in few-view scenarios, 3DGS tends to over-reconstruct available high-frequency (HF) regions, sharply reproducing textures and edges in training views, while losing smooth low-frequency (LF) structures and overall stability of the scene\cite{nguyen2025dwtnerf}. 

We address the few-view 3D reconstruction challenge with our proposed method, a frequency-aware extension of 3DGS that integrates the Discrete Wavelet Transform (DWT) to guide spatial-frequency learning. By decomposing rendered and ground-truth images into multi-scale sub-bands, the model explicitly balances both lowand high frequency information. Two complementary supervision strategies are introduced: a global DWT loss that preserves large-scale consistency and a patch-wise DWT loss that refines local details and fine edges. Together, these components balance between HF and LF regions, yielding sharper, more reliable reconstructions under sparse supervision. Building on this foundation, we extend the framework to handle multispectral data. The proposed MultiSpectral 3DGS jointly reconstructs RGB and Near-Infrared (NIR) views using shared geometry and modality-specific appearance parameters, ensuring spectral and spatial coherence across bands. To support research in this area, we introduce a new multispectral greenhouse dataset along with an open-source few-shot benchmarking package that standardizes sparse-view evaluation protocols.  We validate both the RGB and multispectral versions of LGDWT-GS on standard benchmarks (LLFF~\cite{mildenhall2019llff}, MipNeRF360~\cite{barron2022mip}) and on our new dataset. The main contributions of this work are summarized as follows:
\begin{itemize}
    \item Introduction of joint local and global frequency-domain supervision to improve 3D reconstruction.  
    \item Development of an open-source multispectral greenhouse dataset containing four spectral bands (Red, Green, Red-Edge, NIR).  
    \item Extension of 3DGS to the multispectral domain, enabling consistent cross-spectral reconstruction.  
    \item Establishment of a standardized few-shot 3DGS benchmark and evaluation protocol.  
\end{itemize}

\begin{figure*}[h!]
    \centering
    \includegraphics[width=\linewidth]{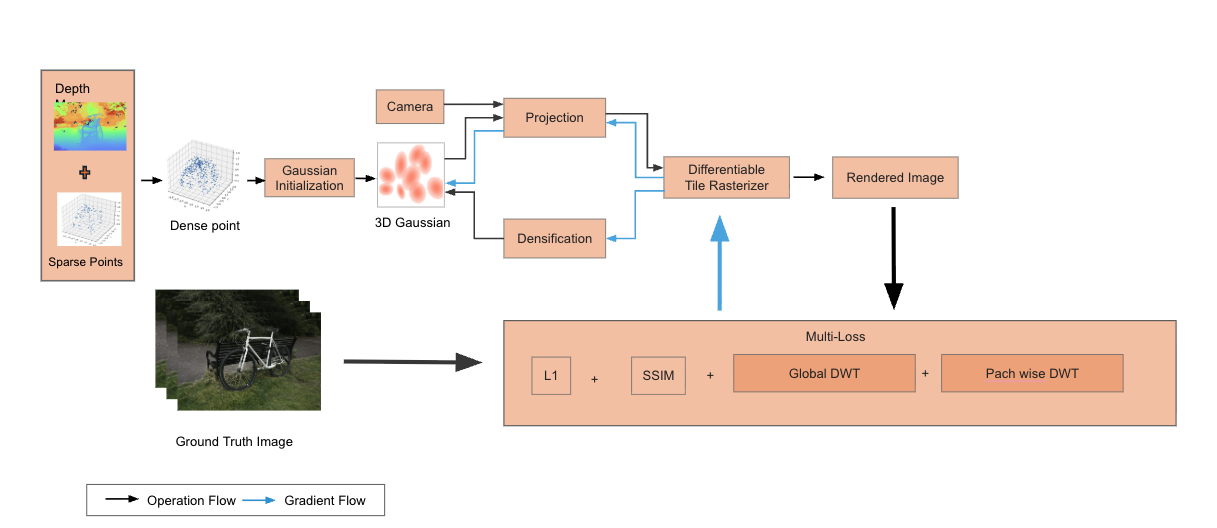}
    \caption{
        Overview of the LGDWT-GS framework.
        The model introduces frequency-domain regularization through global and local DWT losses.
        Combined supervision (L1, SSIM, and global DWT and local DWT) enhances structural stability and textural fidelity.
        Black and blue arrows represent operation and gradient flows, respectively.
    }
    \label{fig:method1}
\end{figure*}
\section{Related Work}

\subsection{Few-Shot Novel View Synthesis}
Building upon the foundational NeRF framework, recent methodologies have emerged to address the challenges of few-shot novel view synthesis. Early approaches in this domain typically rely on strong regularization priors. For instance, RegNeRF~\cite{niemeyer2022regnerf} regularizes geometry by enforcing smoothness constraints on unobserved viewpoints, while FreeNeRF~\cite{yang2023freenerf} employs frequency regularization to prevent HF artifacts during the early stages of training. Other methods, such as SparseNeRF~\cite{wang2023sparsenerf} and DietNeRF~\cite{jain2021putting}, incorporate auxiliary supervision to guide reconstruction. Specifically, SparseNeRF utilizes depth priors, while DietNeRF leverages semantic consistency losses to guide geometry in occluded regions.

Recent advancements have focused on robust geometric adaptation without heavy external priors. FrugalNeRF~\cite{lin2025frugalnerf} introduces a cross-scale sharing scheme to maximize information utility from limited pixels. Furthermore, addressing the practical reality of agricultural and robotic data, methods like SPARF~\cite{truong2023sparf} have extended few-shot capabilities to handle noisy camera poses, jointly refining extrinsic parameters and scene geometry to prevent drift in uncontrolled environments.

\subsection{3DGS and Sparse-View Extensions}

3DGS substantially reduces the training latency of NeRF-based methods through efficient differentiable rasterization. Despite this advantage, 3DGS exhibits a characteristic failure mode in sparse-view or few-shot regimes. In such settings, the optimization process tends to over-reconstruct HF details, particularly edges and fine textures, in the observed training views. Due to limited multiview constraints, this behavior leads to poor generalization across viewpoints and results in degraded global geometric coherence and weakened structural consistency in unobserved regions~\cite{nguyen2025dwtnerf}.

To address these limitations, several extensions have been proposed to improve the robustness of 3DGS under sparse supervision. Methods such as FSGS~\cite{zhu2024fsgs} and PGDGS~\cite{huang2025pgdgs} employ adaptive and progressive densification strategies that incrementally populate underconstrained regions of the scene, thereby reducing geometric sparsity and improving reconstruction stability. Moreover, DNGaussian~\cite{li2024dngaussian} and SCGaussian~\cite{peng2024structure} focus on regularizing scene geometry through explicit depth constraints and structural consistency priors. By suppressing spurious or unstable Gaussian primitives, these approaches enhance geometric plausibility and reduce reconstruction artifacts in sparsely observed areas. While effective, such methods primarily rely on geometric supervision and do not explicitly regulate the spectral distribution of reconstructed content, leaving frequency-domain inconsistencies insufficiently constrained.

\subsection{Frequency-Aware Supervision}

Frequency-domain analysis has been adopted in neural rendering as an effective means to separate global structural information from fine-scale texture. LF components primarily encode smooth geometry and large-scale scene structure, whereas HF components correspond to edges and detailed appearance variations. By explicitly regulating these components, frequency-aware supervision provides a principled mechanism for stabilizing optimization under limited viewpoint coverage.WaveNeRF~\cite{xu2023wavenerf} and DWT-NeRF~\cite{nguyen2025dwtnerf} use wavelet guidance to improve NeRF training. This reduces errors and sharpens edges, especially when there are few camera views. Similarly, DWT-GS~\cite{nguyen2025dwtgs} applies this concept to Gaussian Splatting to remove HF noise. However, these methods generally apply rules to the whole image at once. They fail to distinguish between preserving the main structure and refining fine details.

In addition to frequency-domain approaches, prior work has explored enhancing high-frequency details through spatial-domain supervision. 
For example,  neural edge histogram descriptors (NEHD) \mbox{~\cite{peeples2025histogram,agashe2025neural}} emphasizes edge information using gradient-based objectives to improve classification performance. 
Such methods provide an alternative to our approach, highlighting the distinction between edge-focused spatial supervision and frequency-aware modeling.
Building on these observations, our approach introduces a dual-branch frequency-aware supervision strategy that is directly integrated into the 3DGS rasterization pipeline. A global frequency branch enforces LF consistency to preserve overall geometric structure, while a complementary patch-wise branch selectively refines HF components to recover local details without inducing overfitting.

\subsection{Multispectral and Agricultural 3D Reconstruction}

Multispectral and hyperspectral imaging capture reflectance information beyond the visible spectrum, enabling critical applications in agriculture~\cite{karukayil20253d}, remote sensing~\cite{ziemann2025new,nia2025neighborhood}, and material analysis~\cite{klein2023physics}. By providing wavelength-dependent measurements, these modalities support robust characterization of vegetation health, structural properties, and material composition that cannot be inferred from RGB imagery alone. Several NeRF-based extensions have been proposed to model spectral radiance fields across multiple wavelengths. Methods such as HS-NeRF~\cite{chen2024hyperspectral}, SpectralNeRF~\cite{li2024spectralnerf}, and Spec-NeRF~\cite{li2024spec} explicitly reconstruct spectral reflectance by conditioning radiance fields on wavelength information. More recently, HyperGS~\cite{thirgood2025hypergs} adapts Gaussian Splatting to hyperspectral scenes, demonstrating the feasibility of point-based spectral rendering. However, these approaches generally assume dense multi-view acquisition and rely on high-cost sensing hardware, limiting their applicability in real-world agricultural settings.

In practice, agricultural imaging is often performed under few-view, spectrally heterogeneous conditions, with additional challenges arising from varying illumination, plant self-occlusions, and complex scene geometry. To better reflect these constraints, we introduce a controlled multispectral greenhouse dataset capturing Red, Green, Red Edge, and NIR channels, along with a unified few-shot benchmarking package. This dataset enables systematic evaluation of frequency-aware multispectral 3D reconstruction methods under realistic agricultural imaging conditions.

\section{Method}

We introduce the LGDWT-GS framework, which integrates frequency-domain supervision into the 3DGS  pipeline to improve few-shot 3D reconstruction.  
Then, we extend this framework to a multispectral version that jointly reconstructs RGB and NIR modalities under a shared 3D geometry.
\subsection{LGDWT-GS Framework}

The proposed method extends the original 3DGS by introducing frequency-aware supervision through global and local (patch-wise) DWT losses. 
This integration enhances large-scale structural consistency and fine-grained texture recovery without modifying the differentiable splatting renderer. In few-shot settings, only a small number of sparse points can be reconstructed by COLMAP~\cite{schonberger2016structure} due to limited view overlap. 
To obtain a denser and more stable initialization, depth priors and multi-view stereo reconstructions are used to generate additional pseudo-points, forming a more complete geometric basis for the Gaussian representation. 
Each Gaussian primitive is parameterized by a mean vector, covariance matrix, color, and opacity.

During training, the differentiable rasterizer projects these Gaussians into the image plane, and the rendered outputs are optimized against ground-truth views using a composite loss function, shown in Equation \ref{eq:ltotal}, that jointly balances spatial, perceptual, and frequency domain objectives

\begin{equation}
\mathcal{L}_{\text{total}} =
\mathcal{L}_{\text{L1}} +
\mathcal{L}_{\text{SSIM}} +
\alpha \mathcal{L}_{\text{Global-DWT}} +
\beta \mathcal{L}_{\text{Patch-DWT}}
\label{eq:ltotal}
\end{equation}

\noindent $\mathcal{L}_{\text{L1}}$ ensures pixel-wise fidelity, $\mathcal{L}_{\text{SSIM}}$ enforces structural similarity, and $\mathcal{L}_{\text{Global-DWT}}$ and $\mathcal{L}_{\text{Patch-DWT}}$ regularize global and local frequency components, respectively. 
The weighting factors $\alpha$ and $\beta$ control the relative contributions of global and patch-wise DWT supervision during optimization. The overall proposed method with this loss function is shown in Figure~\ref{fig:method1}.

\subsubsection{Global DWT Supervision}

\begin{figure}[!t]
\centering

\begin{minipage}{0.24\linewidth}
\centering
\includegraphics[width=\linewidth]{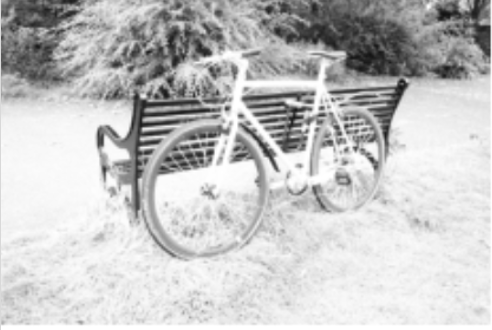}
\\ (a)
\end{minipage}
\hfill
\begin{minipage}{0.24\linewidth}
\centering
\includegraphics[width=\linewidth]{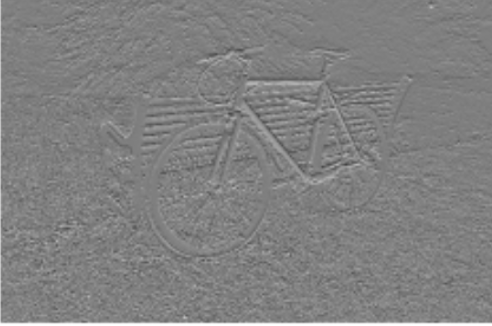}
\\ (b)
\end{minipage}
\hfill
\begin{minipage}{0.24\linewidth}
\centering
\includegraphics[width=\linewidth]{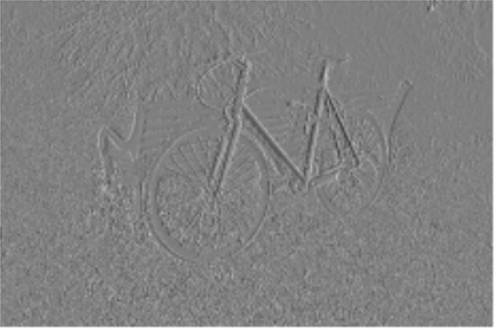}
\\ (c)
\end{minipage}
\hfill
\begin{minipage}{0.24\linewidth}
\centering
\includegraphics[width=\linewidth]{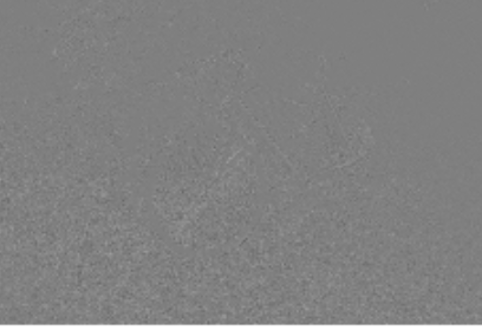}
\\ (d)
\end{minipage}

\caption{Wavelet decomposition of the input image into four subbands:
(a) $I_{\text{LL}}$ (approximation),
(b) $I_{\text{LH}}$ (horizontal detail),
(c) $I_{\text{HL}}$ (vertical detail),
(d) $I_{\text{HH}}$ (diagonal detail).}
\label{fig:wavelet_subbands}
\end{figure}

The global DWT loss enforces large-scale frequency consistency between rendered and ground-truth images, promoting structural stability across views.  
Each image is decomposed using a one-level Haar wavelet transform \mbox{~\cite{cotter2019pytorch_wavelets}} into four frequency subbands. The \text{LL} subband represents the LF approximation of the image and preserves its coarse global structure, the \text{LH} subband captures horizontal HF components corresponding to vertical edges, the \text{HL} subband captures vertical HF components corresponding to horizontal edges, and the \text{HH} subband contains diagonal HF information that highlights fine, corner-like details that are shown in Figure~\ref{fig:wavelet_subbands}.

The global DWT loss term (Equation \ref{eq:global_dwt}) is defined as follows:

\begin{equation}
\mathcal{L}_{\text{Global-DWT}} =
\sum_{s \in \{\text{LL}, \text{LH}, \text{HL}, \text{HH}\}}
w_s \, \|\hat{I}_{s} - I_{s}\|_1,
\label{eq:global_dwt}
\end{equation}

where $w_s$ denotes the frequency weight for each subband.  
To mitigate overfitting to unstable HF regions ~\cite{yang2023freenerf}, the HH component is included but assigned a weight near zero. In our implementation, the HH component is set to zero to avoid introducing unstable high-frequency noise in sparse-view settings \mbox{~\cite{nguyen2025dwtgs}}
This formulation aligns global frequency structures between rendered and reference images, ensuring coherent reconstruction of both LF and mid-frequency content while suppressing noisy HF artifacts.

\subsubsection{Patch-wise DWT Supervision}
Unlike prior DWT-based approaches such as DWT-NeRF and DWT-GS, which apply frequency supervision globally, our method introduces patch-wise frequency supervision to explicitly address spatially localized reconstruction errors. While global DWT enforces overall frequency consistency, it may overlook regions where fine details are under-reconstructed. In contrast, patch-wise supervision enables the model to focus on challenging regions (e.g., thin structures, texture boundaries, or low-frequency-dominated areas lacking high-frequency detail). This localized refinement improves the balance between low- and high-frequency components within each region, resulting in sharper details and more consistent reconstruction, particularly in sparse-view settings where errors are spatially non-uniform.

Specifically, we introduce patch-wise DWT supervision to refine such regions by locally balancing frequency components. The rendered and ground-truth images are divided into non-overlapping patches of fixed size and stride. Each patch is independently analyzed to detect local frequency imbalance, guided by a low-frequency energy ($E_{\text{LF}}$) metric that quantifies the relative dominance of LF versus HF content (Equation~\ref{eq:energy}).

Each image is first decomposed via a one-level Haar wavelet transform:
where $I_{\text{LL}}$ denotes the low frequency (structural) subband and $I_{\text{HF}}$ represents the aggregated high frequency components $(I_{\text{LH}}, I_{\text{HL}}, I_{\text{HH}})$.

The $E_{\text{LF}}$ for each spatial location is defined as:
\begin{equation}
E_{\text{LF}}(x,y) = \frac{\|I_{\text{LL}}(x,y)\|_1}{\|I_{\text{LL}}(x,y)\|_1 + \|I_{\text{HF}}(x,y)\|_1}
\label{eq:energy}
\end{equation}

\begin{figure}[!t]
\centering

\begin{minipage}{0.48\linewidth}
\centering
\includegraphics[width=\linewidth]{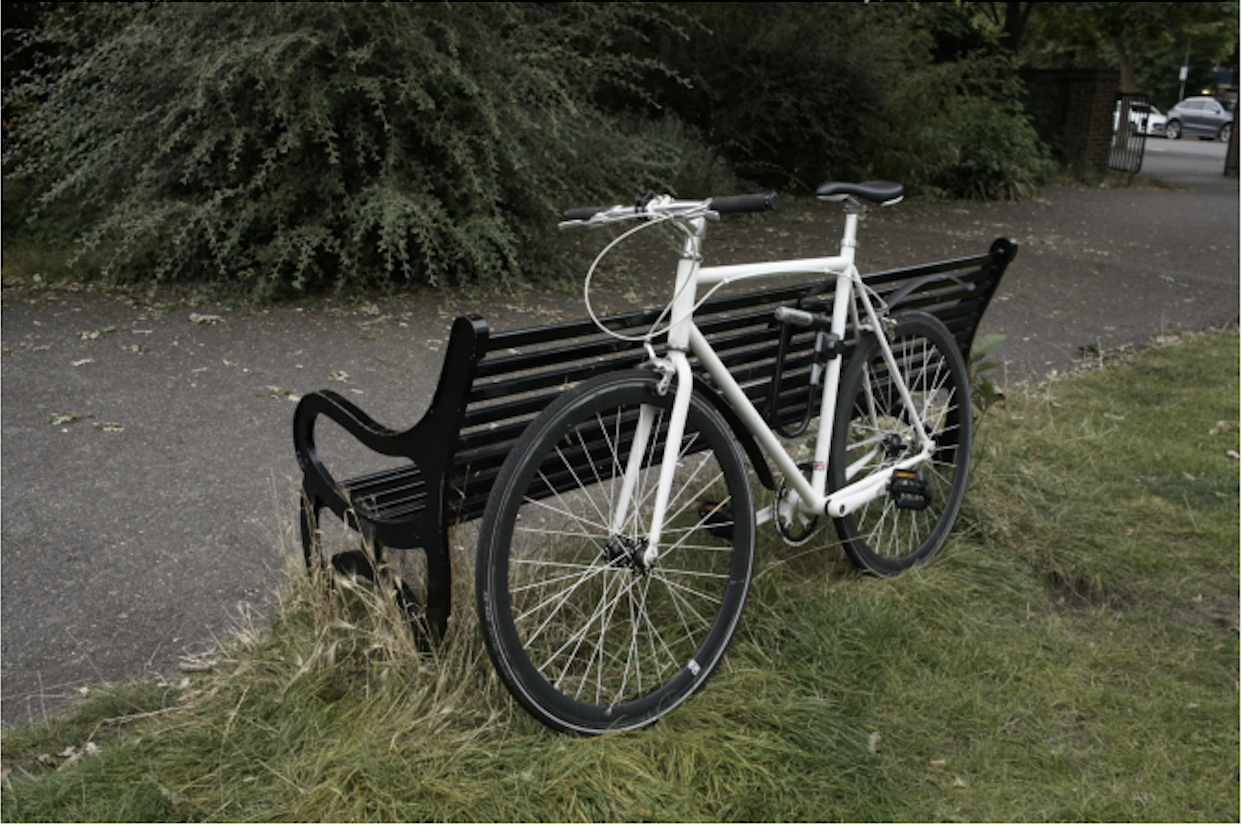}
\\ (a)
\label{fig_lf_gt}
\end{minipage}
\hfill
\begin{minipage}{0.48\linewidth}
\centering
\includegraphics[width=\linewidth]{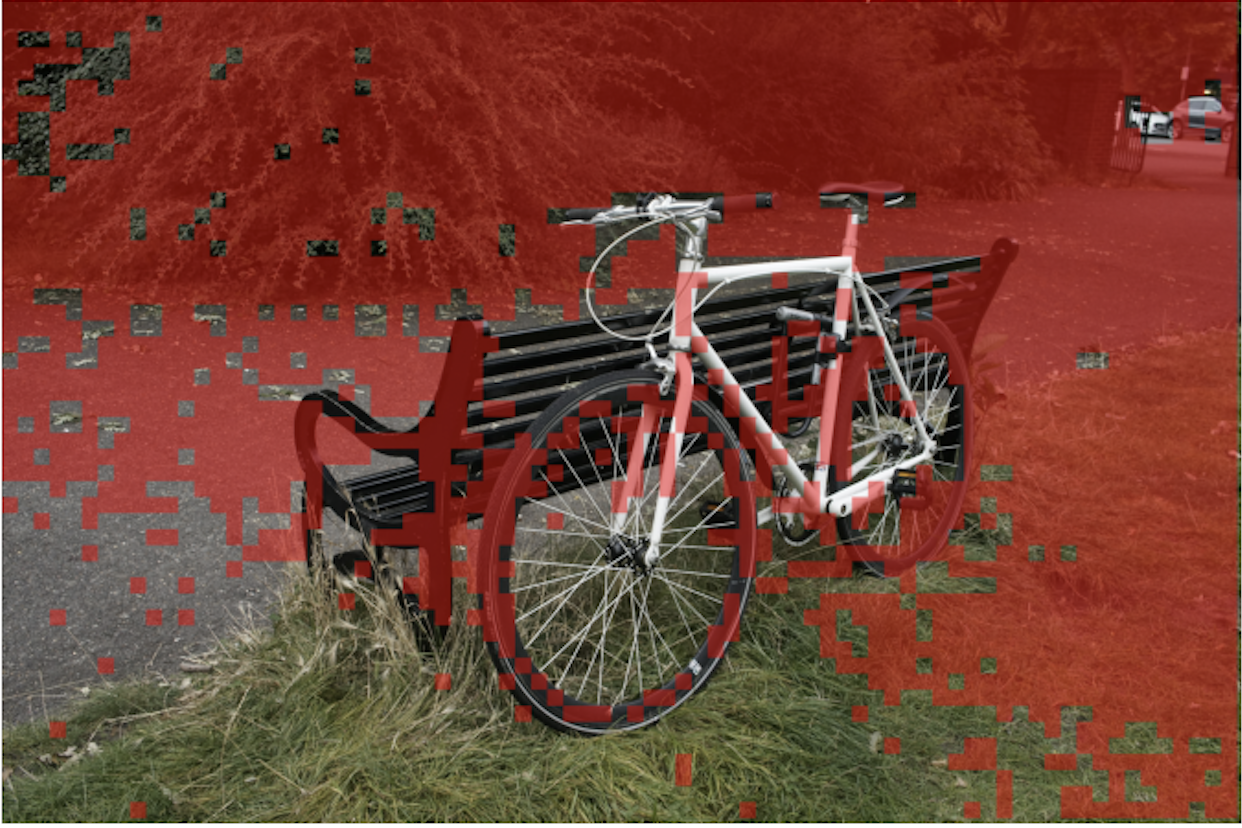}
\\ (b)
\label{fig_lf_map}
\end{minipage}

\caption{
$E_{LF}$ map used for patch selection:
(a) Ground Truth,
(b) $E_{LF}$ Map.
Red regions denote low $E_{LF}$ values, indicating weak LF stability or missing HF details and revealing spatial frequency imbalance in the reconstruction.
}
\label{fig:lfmap}
\end{figure}

\noindent Regions with low $E_{\text{LF}}$ values, defined as pixels whose $E_{\text{LF}}$ falls below a percentile-based threshold, correspond to areas where structural information is weak or where high-frequency details are under-represented within low-frequency regions, as illustrated in Figure~\ref{fig:lfmap}.
These regions are therefore prioritized for localized frequency refinement.

For each selected patch, a one-level Haar wavelet transform is applied, and losses are computed on the high-frequency subbands to enhance local detail reconstruction. The patch-level loss in Equation~\mbox{\ref{eq:patch_dwt}} is defined as:
\begin{equation}
\mathcal{L}_{\text{Patch-DWT}} =
\frac{1}{N_p}
\sum_{p=1}^{N_p}
\sum_{s \in \{\text{LH}, \text{HL}\}}
\left\| \hat{I}^{\,p}_{\,s} - I^{\,p}_{\,s} \right\|_1,
\label{eq:patch_dwt}
\end{equation}

\noindent where $N_p$ denotes the number of selected patches,
$s \in \{\text{LH}, \text{HL}\}$ the corresponding frequency subband,
$\hat{I}^{\,p}_{\text{s}}$ the rendered patch wavelet coefficients,
and $I_{\text{s}}^p$ the corresponding ground-truth patch wavelet coefficients in subband $s$ for patch $p$. This localized supervision improves fine-grained reconstruction by reinforcing HF corrections within LF regions, sharpening object boundaries, and restoring texture details while maintaining global stability.

\subsection{MultiSpectral Extension}

The LGDWT-GS framework is extended to support dual-modality reconstruction, jointly modeling RGB and NIR images under a shared 3D geometry.  
This multispectral formulation introduces cross-spectral supervision, improving both geometric fidelity and spectral alignment across modalities. 
To initialize the geometry, pseudo-RGB images are first generated by combining the Red, Green, and Red-Edge spectral channels.  
These pseudo-RGB images are used as inputs to COLMAP to estimate camera poses and generate the initial sparse point cloud.  
The resulting structure provides reliable geometric alignment across all spectral bands, even in sparse-view conditions.

The RGB and NIR data are stored in separate but spatially aligned folders, sharing identical intrinsic and extrinsic camera parameters.  
This setup ensures pixel-level correspondence between modalities and enables synchronized multispectral loading during training.  
Optional depth priors can also be incorporated to further densify the initialization and stabilize reconstruction in regions with limited multiview overlap. 
Each Gaussian primitive maintains shared geometric parameters while encoding modality-specific color attributes.  
Two-pass differentiable rasterization is then applied to produce the corresponding renderings, allowing both branches to be optimized jointly under a unified geometry while preserving spectral distinctions.

Supervision combines reconstruction objectives from both the RGB channel and the NIR channel:
\begin{equation}
\mathcal{L}_{\text{Multi}} =
\mathcal{L}_{\text{RGB}} + \lambda_{\text{NIR}}\, \mathcal{L}_{\text{NIR}},
\label{eq:multi}
\end{equation}

\noindent where $\lambda_{\text{NIR}}$ controls the relative contribution of the NIR branch.

During densification, new Gaussians are spawned in regions exhibiting high residuals in either spectral channel:
\begin{equation}
\text{mask} = \max(\text{RGB}_{\text{res}}, \text{NIR}_{\text{res}}),
\label{eq:mask}
\end{equation}

\noindent ensuring that both spectral domains selectively guide geometric refinement. 
The use of the $\max$ operator is motivated by the need to capture the strongest reconstruction error across modalities (RGB and NIR) at each spatial location. Since our goal is to trigger densification whenever any modality exhibits a significant error (i.e., under-reconstruction), the $\max$ operation ensures that such regions are not overlooked.

In contrast, a weighted sum may dilute strong errors in one modality when the other modality has low residuals, potentially suppressing important signals for densification. While one could introduce weights to address this, doing so would require either manual tuning or learning additional parameters, increasing model complexity and training overhead. Moreover, learnable weights may introduce instability and reduce the interpretability of the densification criterion. Similarly, threshold-based or union strategies require additional hyperparameters, making the method more sensitive to hyperparameter selection. Overall, the $\max$ operator provides a simple, parameter-free, and robust formulation that prioritizes worst-case reconstruction error across modalities, ensuring consistent refinement of high-error regions in a stable and efficient manner.
This shared-geometry, dual-appearance design promotes cross-spectral coherence and enhances fine-detail reconstruction under sparse-view conditions.

Multispectral reconstruction is inherently a multi-frequency problem, as different spectral bands share low-frequency structural information while exhibiting distinct high-frequency characteristics. 
Without frequency-aware modeling, these cross-spectral differences can lead to inconsistent geometry and unstable optimization. Our DWT-based formulation explicitly decomposes and aligns frequency components across modalities. 
In particular, low-frequency alignment stabilizes shared geometric structure, while separating high-frequency bands allows the model to preserve complementary spectral details (e.g., vegetation patterns in NIR) without forcing them into a common representation. 
This also reduces cross-spectral interference during optimization. Therefore, the benefit of DWT in the multispectral setting goes beyond simply adding supervision. The proposed method enables structured cross-spectral frequency alignment, which is critical for stable and accurate reconstruction.

\begin{figure}[t]
    \centering
    \includegraphics[width=\linewidth]{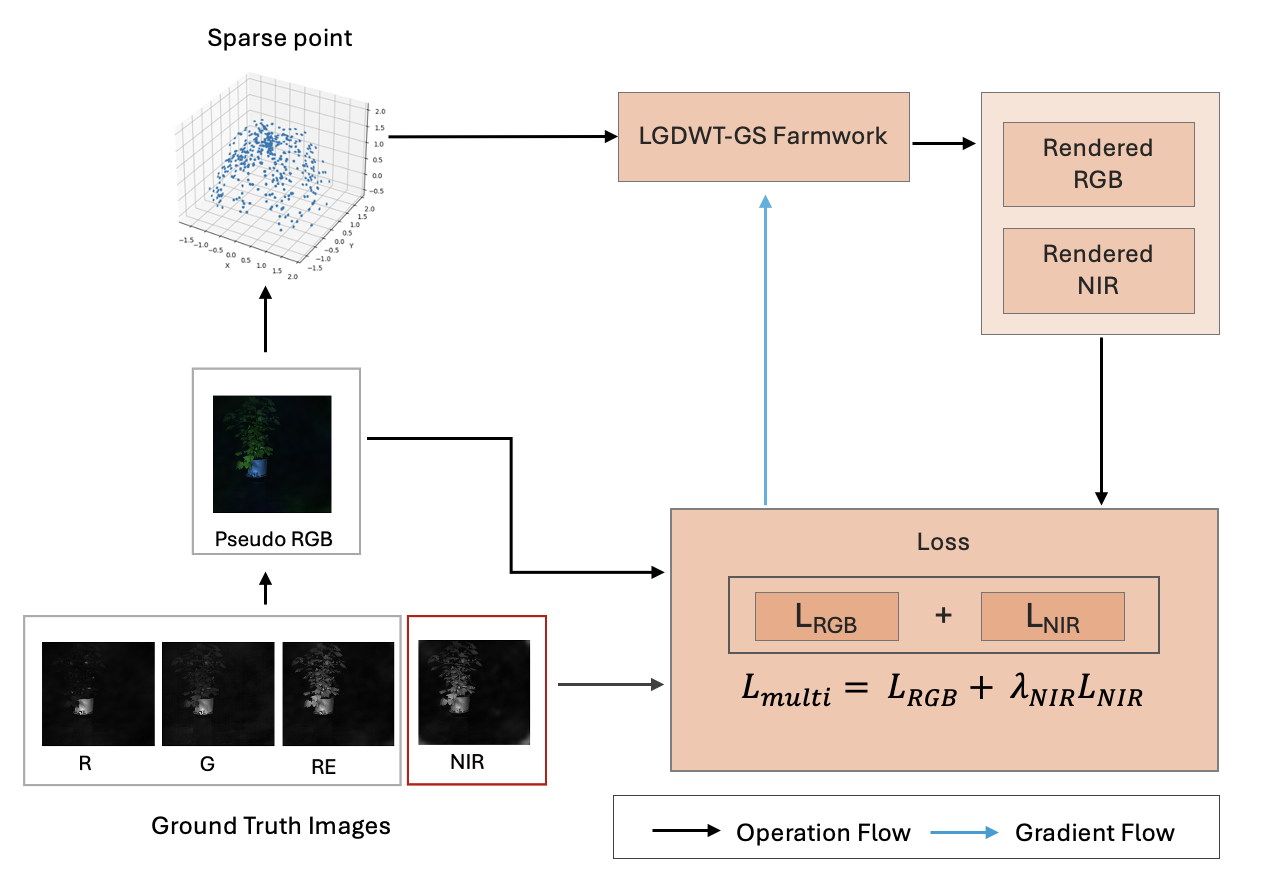}
    \caption{
        Multispectral LGDWT-3DGS framework. 
        Pseudo-RGB images (constructed from Red, Green, and Red-Edge bands) are used for COLMAP-based pose estimation and sparse reconstruction. 
        RGB and NIR chanells are then jointly optimized under a shared geometry using cross-spectral supervision and DWT-based frequency regularization, improving geometric consistency and spectral alignment under sparse-view scenarios.
    }
    \label{fig:method2}
\end{figure}

\subsection{Dataset}
\begin{figure*}[!t]
    \centering
    \includegraphics[width=0.95\textwidth]{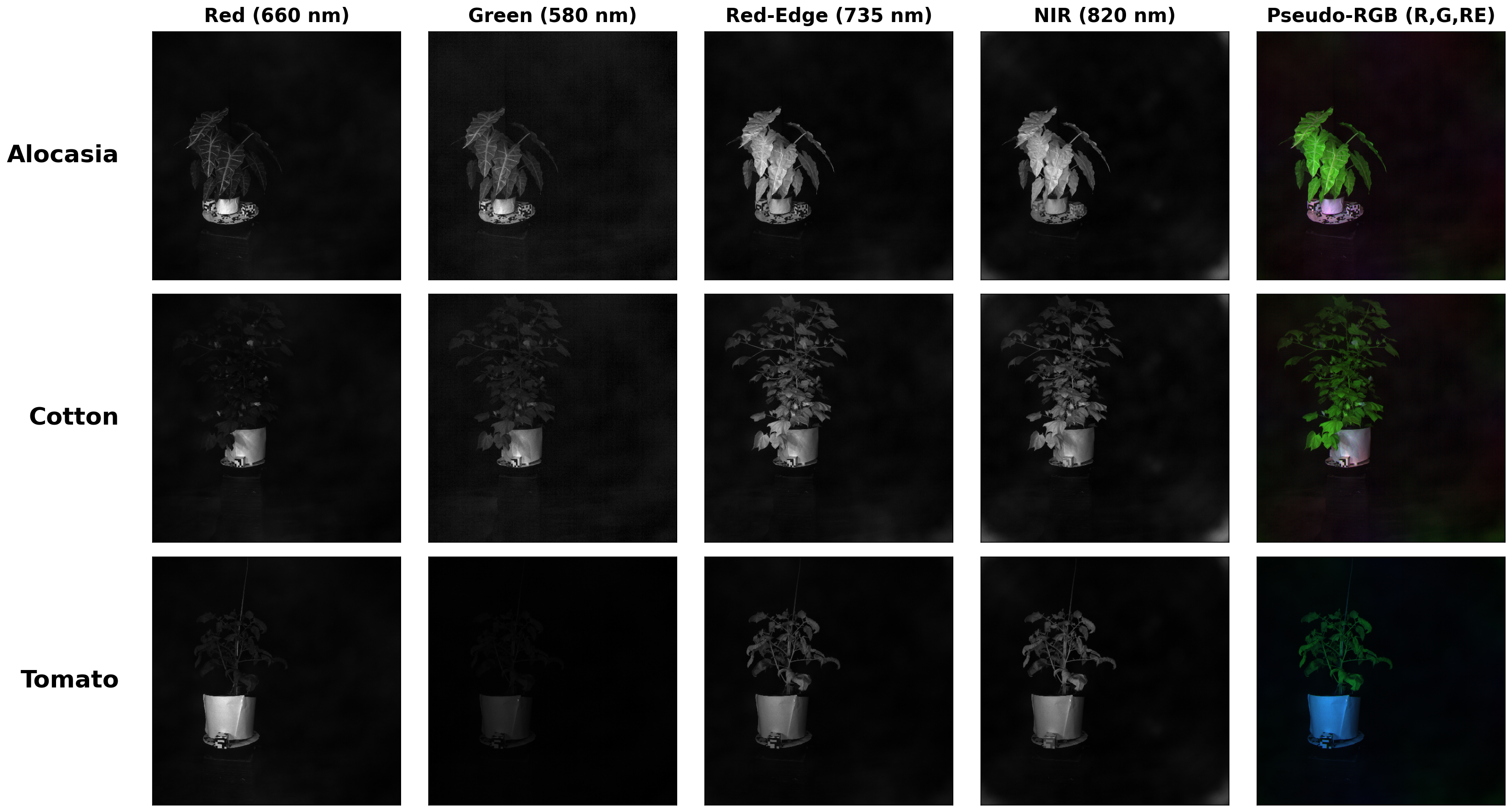}
    \caption{Example spectral channels for three representative plant scenes. Columns correspond to 580~nm (Green), 660~nm (Red), 735~nm (Red Edge), and 820~nm (NIR) bands. The final column shows the pseudo-RGB composite used for COLMAP reconstruction.}
    \label{fig:multispectral_samples}
\end{figure*}

To construct the proposed multispectral greenhouse dataset, we employed the MSIS-AGRI-1-A system: a snapshot multispectral camera equipped with four spectral bands centered at 580~nm (Green), 660~nm (Red), 735~nm (Red Edge), and 820~nm (NIR)  Figure~\ref{fig:multispectral_samples}.  
The camera uses a global shutter CMOS sensor with 4~MP resolution and integrates Anti-X-Talk\texttrademark~technology to minimize spectral leakage between channels, ensuring radiometrically accurate and high-contrast measurements.  
Each capture was synchronized with a four-channel LED illumination module operating at the same wavelengths to maintain consistent spectral lighting across all bands.  
Before data collection, both cameras were calibrated for intrinsic and extrinsic parameters using a checkerboard target to ensure precise geometric alignment.

\begin{figure}[htb]
    \centering
    \includegraphics[width=0.65\linewidth]{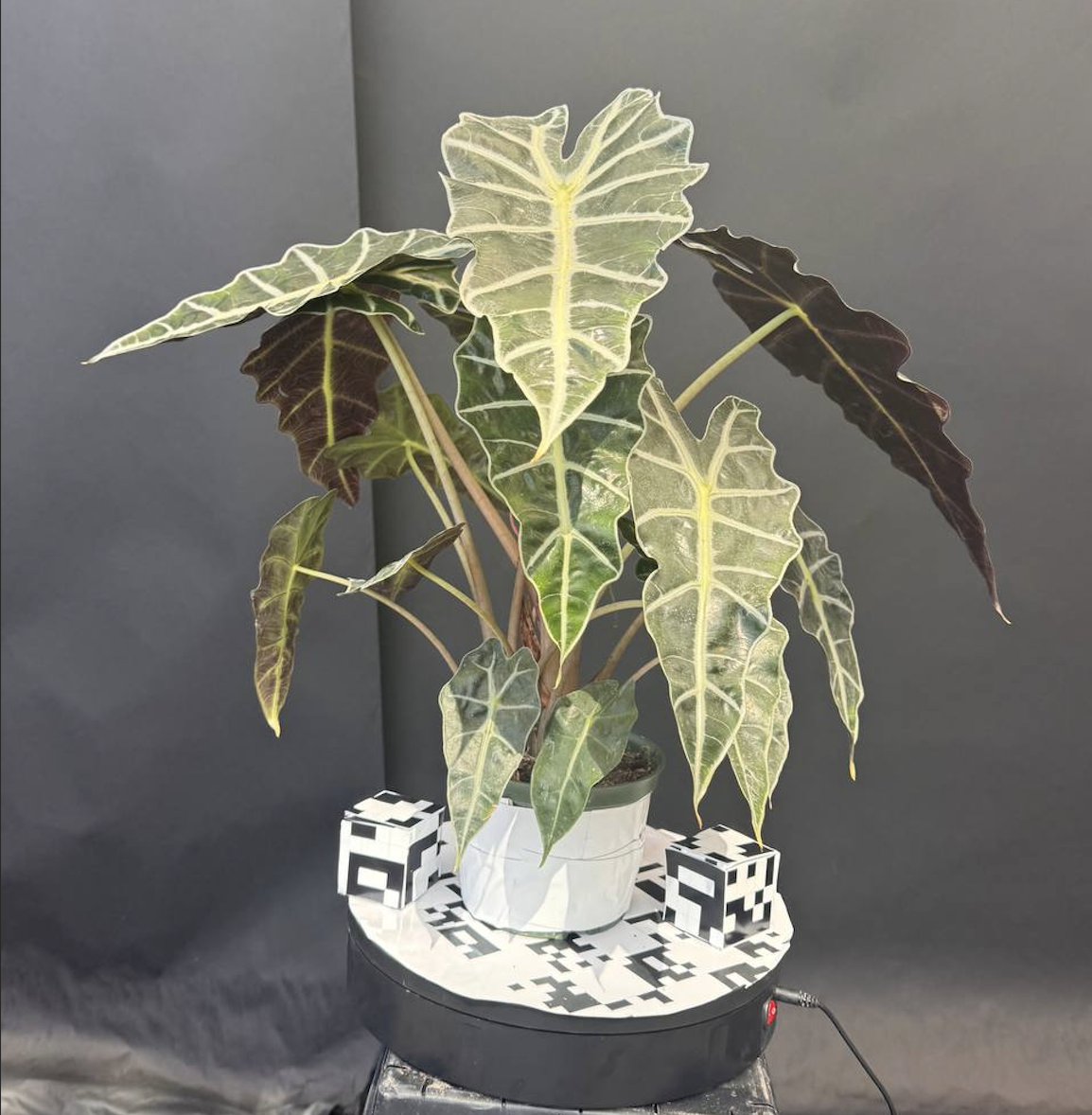}
    \caption{Greenhouse imaging setup. The MSIS-AGRI-1-A camera, LED illumination system, motorized turntable, and cube reference markers used for geometric calibration are shown.}
    \label{fig:env_setup}
\end{figure}

Each scene corresponds to an individual plant species sorghum, tomato, alocasia, cotton, and grape captured inside a controlled greenhouse imaging station Figure~\ref{fig:env_setup}.  
The setup includes a motorized turntable, a uniform black background, and a multispectral LED illumination system designed to ensure radiometric consistency and suppress ambient reflections.  
Two identical cameras were placed on opposite sides of the turntable, providing dual viewpoints.  
For each fixed horizontal position, the cameras were vertically translated across four discrete heights to capture multiple canopy layers.  
Each plant was imaged at ten rotational steps (36° increments) using the motorized turntable, with two reference cube markers attached for rotation tracking and geometric calibration.  
Each acquisition session generated approximately 80–100 multispectral frames per scene, yielding almost 500 spatially aligned spectral images across all plant species.

\begin{figure}[htb]
    \centering
    \includegraphics[width=\linewidth]{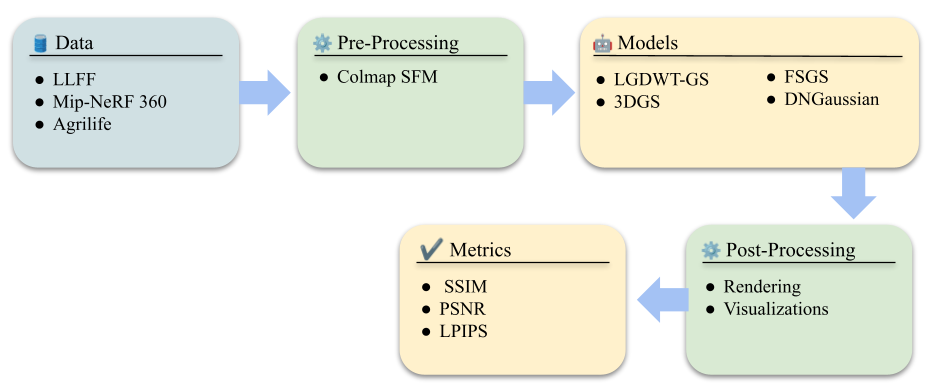}
    \caption{Few-shot 3DGS benchmarking tool overview. The overall framework and figure is adapted from Anomalib~\cite{akcay2022anomalib}.}
    \label{fig:benchmark}
\end{figure}

For each acquisition, four spectral bands were recorded simultaneously, forming a co-registered multispectral image cube $\{I_{R}, I_{G}, I_{\text{RE}}, I_{\text{NIR}}\}$.  
A pseudo-RGB composite was generated from the Red, Green, and Red Edge channels to facilitate Structure-from-Motion (SfM) reconstruction using COLMAP.  
Reconstructed camera poses, sparse point clouds, and scene bounds were used as geometric initialization for multispectral 3DGS training.  
This preprocessing ensures spatial alignment between spectral modalities and enables consistent cross-spectral supervision during reconstruction.

\section{Experiments}

\subsection{Implementation Details}

In our implementation, we use a one-level discrete wavelet transform (DWT) with fixed weighting coefficients $\alpha = 1.0$ and $\beta = 0.5$ for global and patch-wise frequency supervision, respectively. While these core hyperparameters remain consistent across all experiments, certain frequency-related weights are adjusted per scene to account for variations in texture complexity, noise levels, and spectral characteristics. For stability in sparse-view settings, the high-frequency HH subband is assigned a weight of zero to suppress unstable diagonal noise components.

For patch-wise supervision, images are divided into non-overlapping patches of size $128 \times 128$ with a stride of 128. Patch selection is guided by the ELF metric, where regions corresponding to the lowest $20\%$ of the ELF distribution are prioritized for refinement. In the multispectral setting, we set the weighting factor $\lambda_{\text{NIR}} = 1.0$ to balance RGB and NIR supervision. All implementation details, including per-scene configurations and frequency weighting strategies, are provided in our public repository and all models are trained on a single NVIDIA A100 GPU.

\subsection{Benchmarking}

 \begin{figure}[h!]
    \centering
    \includegraphics[width=\linewidth]{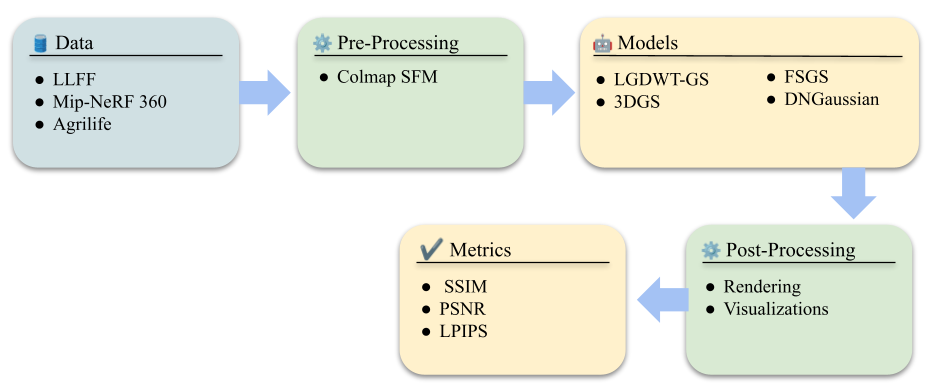}
    \caption{Few-shot 3DGS benchmarking tool overview. The overall framework and figure is adapted from Anomalib~\cite{akcay2022anomalib}.}
    \label{fig:benchmark}
\end{figure}

We introduce an end-to-end benchmarking pipeline for few-shot 3D reconstruction that unifies data processing, training, and evaluation across multiple Gaussian Splatting baselines. While existing frameworks such as Nerfstudio~\cite{tancik2023nerfstudio} simplify NeRF-style experimentation, they do not yet provide native support for few-shot Gaussian Splatting workflows or standardized evaluation under sparse-view conditions.

Our system ingests common multi-view datasets and executes a unified SfM stage using COLMAP to ensure consistent camera poses and sparse geometry across all experiments. On top of this shared foundation, the pipeline supports the training of multiple few-shot Gaussian Splatting variants, including 3DGS~\cite{kerbl20233d}, FSGS~\cite{zhu2024fsgs}, and DNGaussian~\cite{li2024dngaussian}, using a common configuration and logging interface.

The benchmarking package is fully modular, allowing new datasets and new Gaussian Splatting methods to be registered through a unified API without re-engineering the environment or modifying existing pipelines. This design enables a one-install, many-model workflow with a fixed data layout, standardized metrics, and consistent evaluation protocols. As a result, comparisons across models, scenes, and random seeds become fair, repeatable, and reproducible, minimizing environment-dependent variability and facilitating reliable analysis of few-shot 3D reconstruction performance.

\subsection{Quantitative Results}

We evaluate the proposed framework on standard benchmarks including LLFF~\cite{mildenhall2019llff}, MipNeRF360, and our controlled greenhouse multispectral dataset. The evaluation is designed to assess reconstruction quality under varying levels of view sparsity and spectral diversity. Despite its efficiency, the proposed LGDWT-GS pipeline converges in under three minutes per scene on an NVIDIA A100 GPU, and wall-clock time may vary on different hardware configurations. As shown in Table\mbox{~\ref{tab:ablation}}, the Global+Local DWT design introduces a modest overhead of pproximately 38\% ($\sim$166s) compared to the single-branch 
DWT baseline ($\sim$120s), which arises from the additional 
patch selection and wavelet decomposition in the local branch. 
This overhead remains constant with respect to scene complexity 
and does not affect convergence stability.

Although the proposed framework supports arbitrary numbers of input views, we report results under representative configurations that are commonly adopted in prior work. Specifically, we evaluate on LLFF using three input views, which is a standard few-shot setting. For MipNeRF360, we report results using 24 input views, which is widely treated as a few-shot configuration due to the dataset’s complex geometry and wide viewpoint variation. For the greenhouse dataset, we adopt a fixed few-shot setting with ten views per plant, reflecting realistic agricultural capture conditions.

Table~\ref{tab:llff_comparison} shows that LGDWT-GS improves performance significantly on the three-view LLFF dataset. These results confirm that frequency supervision prevents the overfitting and structural errors often seen in sparse-view training. LGDWT-GS also consistently outperforms DNGaussian, suggesting that frequency control offers benefits that geometry modeling alone cannot provide. Additionally, our method achieves higher scores than NeRF-based methods like RegNeRF and FreeNeRF. This proves that enforcing consistency in the frequency domain preserves both global structure and fine details, even with very few views.

Also we observe that performance degradation under global-only DWT supervision occurs consistently across scenes, as evidenced by the ablation results in Table ~\mbox{\ref{tab:ablation}}, particularly in regions containing fine structures and localized high-frequency details. While we do not attribute this behavior to a single definitive cause, it is consistent with insufficient reconstruction of localized high-frequency content, especially in regions where such details are embedded within smoother low-frequency structures.

This interpretation is further supported by our ablation study (Table~\mbox{\ref{tab:ablation}}), where global-only DWT supervision yields lower performance, whereas the proposed Global + Local DWT formulation consistently improves PSNR, SSIM, and LPIPS. Since the only difference between these configurations is the inclusion of localized frequency supervision, this strongly suggests that the observed performance gap arises from limitations in capturing local high-frequency details under global-only supervision.

Table~\ref{tab:mipnerf_comparison} presents results on MipNeRF360 using 24 input views. Even with more images, this dataset remains difficult due to complex scenes and wide angles. Under these conditions, LGDWT-GS achieves the highest overall scores. The strong performance against DNGaussian shows that frequency supervision is effective beyond just sparse data. It ensures stable reconstruction across both limited-view and complex multi-view scenarios.

\begin{table}[h!]
\centering
\caption{Comparison on LLFF (3-View). Metrics include PSNR (dB) $\uparrow$, SSIM $\uparrow$, and LPIPS $\downarrow$.}
\resizebox{0.7\linewidth}{!}{
\begin{tabular}{lccc}
\hline
Method & PSNR $\uparrow$ & SSIM $\uparrow$ & LPIPS $\downarrow$ \\
\hline
Mip-NeRF360 & 15.22 & 0.351 & 0.540 \\
DietNeRF    & 13.86 & 0.305 & 0.578 \\
RegNeRF     & 18.66 & 0.535 & 0.411 \\
FreeNeRF    & 19.13 & 0.562 & 0.384 \\
SparseNeRF  & 19.07 & 0.564 & 0.392 \\
3DGS        & 16.94 & 0.488 & 0.402 \\
DNGaussian  & 19.73 & 0.669 & 0.301 \\
LGDWT-GS (ours) & \textbf{20.46} & \textbf{0.726} & \textbf{0.279} \\
\hline
\end{tabular}}
\label{tab:llff_comparison}
\end{table}

\begin{table}[h!]
\centering
\caption{Comparison on MipNeRF-360 (24-View). Metrics include PSNR (dB) $\uparrow$, SSIM $\uparrow$, and LPIPS $\downarrow$.}
\resizebox{0.7\linewidth}{!}{
\begin{tabular}{lccc}
\hline
Method & PSNR $\uparrow$ & SSIM $\uparrow$ & LPIPS $\downarrow$ \\
\hline
Mip-NeRF360 & 19.78 & 0.530 & 0.431 \\
DietNeRF    & 19.11 & 0.482 & 0.452 \\
RegNeRF     & 20.55 & 0.546 & 0.398 \\
FreeNeRF    & 21.04 & 0.587 & 0.377 \\
SparseNeRF  & 21.13 & 0.600 & 0.389 \\
3DGS        & 19.93 & 0.588 & 0.401 \\
DNGaussian  & 22.13 & 0.676 & 0.301 \\
LGDWT-GS (ours) & \textbf{22.41} & \textbf{0.692} & \textbf{0.298} \\
\hline
\end{tabular}}
\label{tab:mipnerf_comparison}
\end{table}

\begin{figure}[!t]
  \centering
  \includegraphics[width=0.92\columnwidth]{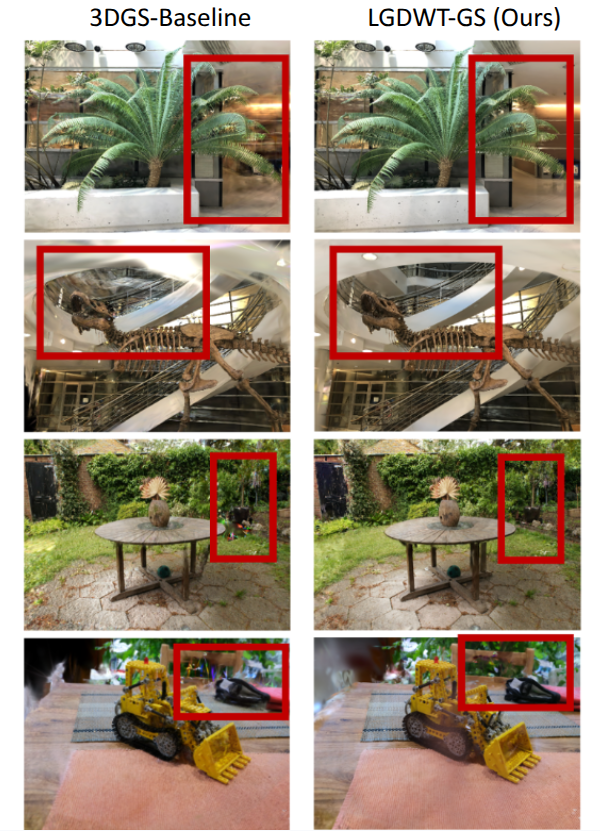}
  \caption{Qualitative comparison between baseline 3DGS and our LGDWT-GS on LLFF and MipNeRF360. Red boxes highlight regions with better preservation of fine details (foliage, edges, thin structures).}
  \label{fig:qualitative}
\end{figure}


To validate LGDWT-GS in a realistic agricultural setting, we evaluated it on our custom multispectral greenhouse dataset. Each plant was captured from ten viewpoints using synchronized RGB and NIR cameras. Table~\ref{tab:multispectral_results} compares the performance against single-channel and standard multispectral baselines. Notably, unlike the LLFF and MipNeRF 360 experiments where high-frequency subbands were excluded, here we explicitly utilized high-frequency regularization. Since the background is irrelevant for this application, we tuned the parameters to focus on the HF components of the plant structure. This targeted approach allows the model to recover fine details, such as leaf veins, without being constrained by background noise. Consequently, LGDWT-GS consistently achieves the highest reconstruction quality across all scenes. In addition, the performance gains arise from three complementary factors. 
First, enforcing a shared 3D geometry across RGB and NIR modalities provides stronger geometric constraints, particularly in sparse-view scenarios where single-modality reconstruction is underconstrained. 
Second, complementary spectral information from the NIR channel captures reflectance properties not present in RGB, especially for vegetation, providing additional structural cues such as leaf boundaries and internal patterns. 
Third, joint optimization introduces cross-spectral supervision, where the combined loss acts as a regularizer that improves stability and reduces overfitting. Together, these factors explain the consistent improvements observed in Table~\mbox{\ref{tab:multispectral_results}}, demonstrating that multispectral supervision enhances both structural consistency and reconstruction fidelity.

However, we observe that the Single + DWT setting slightly underperforms the Single baseline in certain scenes (e.g., Houseplant and Grape). 
Here, “Single” refers to processing each spectral modality independently, rather than jointly exploiting cross-spectral information under a shared geometry. 
In this setting, DWT regularization is applied to each modality separately, without the benefit of complementary supervision from other spectral bands. 
As a result, frequency regularization may over-constrain the reconstruction~\mbox{\cite{nguyen2025dwtnerf}} in scenes with thin structures or complex textures, leading to slight over-smoothing and reduced detail preservation. 
This limitation is alleviated in the multispectral setting, where joint optimization across RGB and NIR provides additional structural cues and more stable supervision. This effect is reflected in Table ~\mbox{\ref{tab:multispectral_results}}, where the Single + DWT configuration shows a slight drop in average PSNR compared to the Single baseline, while Multispectral + DWT achieves the best overall performance.


\begin{table*}[t]
\centering
\caption{Quantitative results on the multispectral greenhouse dataset.}
\label{tab:multispectral_results}
\resizebox{\textwidth}{!}{
\begin{tabular}{|c|ccc|ccc|ccc|ccc|}
\hline
Scene &
\multicolumn{3}{c|}{\textbf{Single}} &
\multicolumn{3}{c|}{\textbf{Single + DWT}} &
\multicolumn{3}{c|}{\textbf{Multispectral}} &
\multicolumn{3}{c|}{\textbf{Multispectral + DWT}} \\
\cline{2-13}
& PSNR↑ & SSIM↑ & LPIPS↓ 
& PSNR↑ & SSIM↑ & LPIPS↓ 
& PSNR↑ & SSIM↑ & LPIPS↓ 
& PSNR↑ & SSIM↑ & LPIPS↓ \\
\hline
Cotton     & 28.74 & 0.820 & 0.422 & 28.96 & 0.829 & 0.424 & 30.55 & 0.873 & \textbf{0.256} & \textbf{30.68} & \textbf{0.874} & 0.258 \\
Grape      & 29.93 & 0.888 & 0.361 & 29.28 & 0.843 & 0.362 & 30.62 & \textbf{0.926} & 0.176 & \textbf{31.01} & 0.925 & \textbf{0.175} \\
Sorghum    & 28.06 & 0.738 & 0.555 & 28.24 & 0.741 & 0.555 & 30.57 & 0.890 & 0.402 & \textbf{31.08} & \textbf{0.894} & \textbf{0.399} \\
Tomato     & 26.65 & 0.788 & 0.357 & 27.08 & 0.802 & 0.351 & 29.33 & 0.885 & \textbf{0.212} & \textbf{29.59} & \textbf{0.892} & 0.213 \\
Houseplant & 29.36 & 0.831 & 0.417 & 28.43 & 0.799 & 0.421 & 29.24 &\textbf{ 0.875} & \textbf{0.245} & \textbf{29.43} & \textbf{0.875} & 0.247 \\
\hline
\textbf{Average} & 28.55 & 0.813 & 0.422 & 28.39 & 0.802 & 0.422 & 30.06 & 0.890 & 0.258 & \textbf{30.51} & \textbf{0.892} & \textbf{0.258} \\
\hline
\end{tabular} }
\end{table*}


\subsection{Qualitative Analysis}

Figure~\ref{fig:qualitative} presents qualitative comparisons between baseline 3DGS and our LGDWT-GS on standard benchmarks.  
Frequency-aware supervision improves detail preservation, notably in foliage, edges, and thin structures.  
Under sparse views, standard 3DGS often exhibits texture blurring or missing details, while LGDWT-GS retains spectral and spatial coherence.  
This qualitative trend is consistent across LLFF and MipNeRF360 datasets.

\begin{figure*}[!t]
\centering

\begin{minipage}{0.18\textwidth}
\centering
\includegraphics[width=\linewidth]{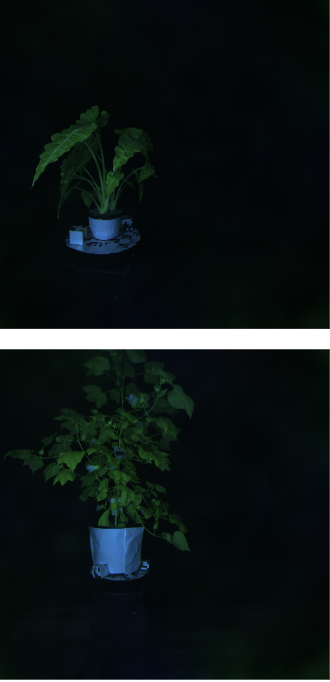}
\\ (a)
\label{fig_gt}
\end{minipage}
\hfill
\begin{minipage}{0.18\textwidth}
\centering
\includegraphics[width=\linewidth]{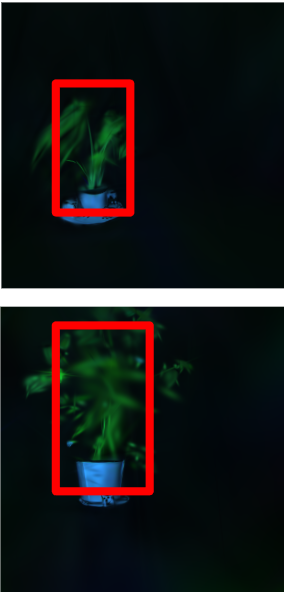}
\\ (b)
\label{fig_sc}
\end{minipage}
\hfill
\begin{minipage}{0.18\textwidth}
\centering
\includegraphics[width=\linewidth]{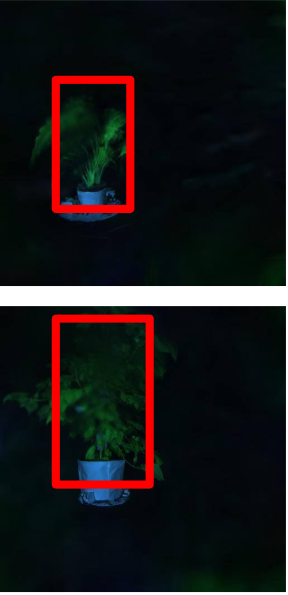}
\\ (c)
\label{fig_sc_dwt}
\end{minipage}
\hfill
\begin{minipage}{0.18\textwidth}
\centering
\includegraphics[width=\linewidth]{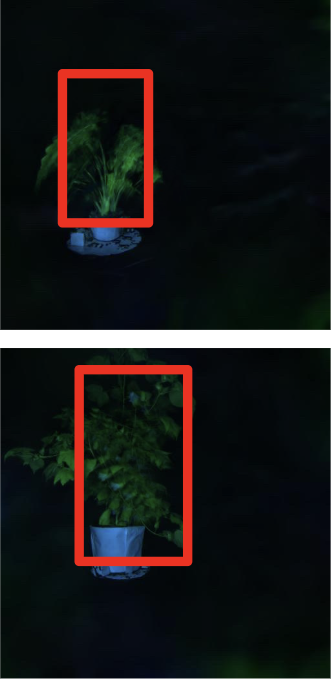}
\\ (d)
\label{fig_ms}
\end{minipage}
\hfill
\begin{minipage}{0.18\textwidth}
\centering
\includegraphics[width=\linewidth]{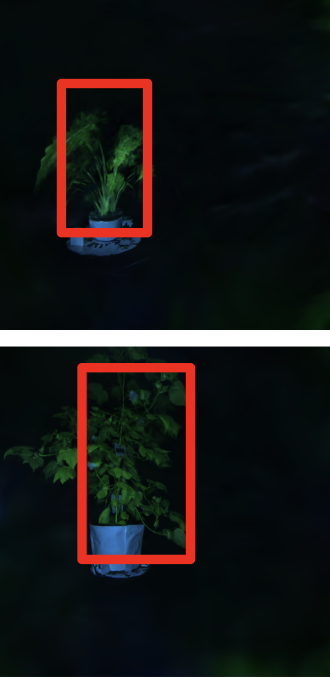}
\\ (e)
\label{fig_ms_dwt}
\end{minipage}

\caption{Comparison of reconstruction outputs across input configurations:
(a) Ground Truth,
(b) Single-channel,
(c) Single-channel + DWT,
(d) Multispectral (no DWT),
(e) Multispectral + DWT.}
\label{fig:multi_compare}
\end{figure*}

Figure~\ref{fig:multi_compare} presents a qualitative comparison on the multispectral greenhouse dataset to analyze the respective contributions of spectral diversity and frequency-domain supervision. We evaluate four reconstruction settings. The single-channel baseline exhibits over-smoothed textures and distorted edges due to limited spectral information. Incorporating DWT improves edge sharpness but does not fully resolve geometric inconsistencies. In contrast, multispectral training enhances global structural stability, though fine details remain blurred. By combining both components, LGDWT-GS achieves the best overall performance, producing sharp and coherent reconstructions with well-preserved leaf boundaries and minimal artifacts. These results demonstrate that spectral diversity promotes geometric stability, while frequency supervision effectively recovers HF details. While LGDWT-GS consistently improves reconstruction quality, we observe certain limitations. 
In information-limited settings, such as single-modality reconstruction, DWT regularization may degrade performance due to over-regularization. 
Additionally, in highly complex scenes (e.g., MipNeRF360), the method may exhibit local inconsistencies despite improving overall reconstruction quality. 
These observations suggest that the effectiveness of frequency-aware supervision depends on the availability and reliability of structural information.

\subsection{Ablation Study}
\label{sec:ablation}

To isolate the contribution of frequency-domain supervision, we conducted a systematic component analysis on the 3-view LLFF configuration (Table~\mbox{\ref{tab:ablation}}). We  include NEHD as a representative edge-focused baseline that enhances high-frequency details through spatial-domain gradient supervision, providing a complementary comparison to our frequency-domain formulation.
 NEHD uses Sobel kernels~\cite{sobel19683x3} to extract edge responses and aggregates them using differentiable histograms of gradient magnitudes to align the edge distributions of rendered and ground-truth images, thereby improving texture representations. During optimization, NEHD explicitly focuses the rendering loss on high-frequency (HF) edge regions. By heavily weighting these fine-detail boundaries, NEHD forces the model to prioritize sharp discontinuities, achieving a PSNR of 19.99~dB. However, this aggressive focus on edges introduces a trade-off: while it sharpens high-contrast boundaries, it often treats subtle surface textures (e.g., leaf veins) as noise or fails to capture them when they lack strong gradient magnitude. This results in reconstructions with sharp outlines but over-smoothed interior regions.

In contrast, LGDWT-GS employs a DWT-based loss to supervise the full frequency spectrum. Instead of relying solely on edge responses, DWT decomposes the signal into distinct sub-bands, effectively decoupling global structure from fine texture. This enables the model to enforce structural consistency through low-frequency (LF) components while simultaneously recovering detailed textures via HF bands, even in regions with weak spatial gradients. Furthermore, in sparse-view settings, DWT acts as a soft regularizer, suppressing HF artifacts in empty space while preserving meaningful signals in textured regions.

To further approximate prior global DWT-based methods such as DWT-GS, we introduce an additional baseline (“sparsepoint + DWT Loss”) in Table~\mbox{\ref{tab:ablation}}. In this setting, the model is initialized using sparse point clouds and trained with global DWT supervision only. This configuration mimics the combination of weaker geometric initialization and global frequency regularization used in prior work. In all other configurations in the ablation study, we use dense point initialization to ensure a strong and consistent geometric prior, isolating the effect of frequency-domain supervision.

As shown in Table~\mbox{\ref{tab:ablation}},this setting results in significantly lower performance, indicating that global-only frequency supervision combined with sparse initialization is insufficient for accurate reconstruction. In contrast, our method benefits from both dense initialization and patch-wise frequency refinement, which together enable more stable geometry and improved detail recovery. We note that exact reproduction of DWT-GS and PGDGS is limited by the lack of publicly available implementations and incomplete reporting of training configurations. Therefore, this baseline provides the closest controlled comparison within our framework.

The component analysis further highlights the contribution of individual design choices. Depth regularization improves PSNR by 0.11~dB by stabilizing geometry and preventing Gaussians from drifting toward the camera, a common failure mode in sparse SfM initialization. We further analyze two extended variants to better understand the role of frequency supervision depth and training strategy. DWT Staging introduces a progressive training scheme in which frequency supervision is not applied at the beginning of training; instead, the model is first optimized without DWT, then the low-frequency (LL) component is introduced, followed by the addition of higher-frequency components (LH, HL) at later stages. This design allows the model to first establish stable coarse geometry before refining fine details, reducing the risk of early overfitting to noisy high-frequency signals.

Two-Level DWT extends the frequency decomposition hierarchy by applying a second level of wavelet transform, enabling multi-scale frequency supervision (e.g., LL$_2$, LH$_2$, HL$_2$). This provides finer-grained control over hierarchical frequency components and improves the model's ability to capture detailed structures, but also increases computational complexity and sensitivity to noise. As shown in Table \mbox{\ref{tab:ablation}}, both variants improve performance compared to single-level DWT, confirming the benefit of deeper or staged frequency supervision. However, the best overall results are achieved by combining global structural constraints with patch-wise local refinement, which effectively balances stability and detail reconstruction.

Finally, combining global structural constraints with patch-based local refinement yields the most balanced performance, achieving the highest SSIM of 0.726 and an LPIPS of 0.279.

\begin{table}[!t]
\caption{Ablation Study on LLFF (3-View) Scenes}
\label{tab:ablation}
\centering
\begin{tabular}{lcccc}
\hline
Configuration        & PSNR (dB)$\uparrow$ & SSIM$\uparrow$ & LPIPS$\downarrow$ & Time (s) \\ \hline

NEHD Loss            & 19.99               & 0.755          & 0.268             & $\sim$360 \\
Sparsepoint + DWT Loss  & 19.09              & 0.667       & 0.427               &
$\sim$120 \\
DWT Loss             & 19.92               & 0.680          & 0.314             & $\sim$120 \\
DWT + Depth Reg.     & 20.03               & 0.683          & 0.304             & $\sim$126 \\
DWT Staging          & 20.08               & 0.720          & 0.298             & $\sim$120 \\
Two-Level DWT        & 20.28               & 0.726          & 0.297             & $\sim$166 \\
Global + Local DWT   & 20.46               & 0.726          & 0.279             & $\sim$166 \\ \hline
\end{tabular}
\end{table}


We also analyze the sensitivity of the $E_{LF}$ threshold used for patch selection. 
Specifically, we vary the threshold across \{10\%, 20\%, 30\%, 50\%\} and report the results in Table \mbox{~\ref{tab:elf_threshold}}. We observe that using a lower threshold (10\%) leads to a noticeable performance drop, as too few regions are selected for local refinement, limiting the effectiveness of patch-wise supervision. 
Increasing the threshold to 30\% yields performance comparable to 20\%, with only a slight degradation due to the inclusion of less informative regions. 
However, when the threshold is increased to 50\%, performance degrades significantly, as a large portion of the image is selected, reducing the selectivity of the supervision and introducing noisy or redundant updates.Overall, 20\% provides the best balance between selectivity and coverage, enabling effective localization of under-reconstructed regions while maintaining stable performance.

\begin{table}[!t]
\centering
\caption{$E_{LF}$ Threshold Sensitivity Analysis on 3-view LLFF.}
\label{tab:elf_threshold}
\begin{tabular}{c|ccc}
\hline
Threshold & PSNR $\uparrow$ & SSIM $\uparrow$ & LPIPS $\downarrow$ \\
\hline
10\% &20.24& 0.714 &  0.287  \\
20\% &20.46&  0.727& 0.279   \\
30\% &20.39& 0.727 &  0.281  \\
50\% &20.31&  0.723&  0.287 \\
\hline
\end{tabular}
\end{table}

\section{Conclusion}
\label{sec:conclusion}

We introduced LGDWT-GS, a frequency-aware extension of 3DGS designed for few-shot 3D reconstruction. By integrating global and patch-wise DWT supervision, our method captures multiscale structural and textural information, mitigating the over-reconstruction tendencies of sparse-view setups. The model preserves both large-scale consistency and fine-grained detail, achieving superior stability and perceptual quality compared to strong baselines such as 3DGS, SparseNeRF, and PGDGS across LLFF, MipNeRF360, and our new greenhouse datasets. Beyond the RGB domain, we extended the framework to a multispectral setting that jointly reconstructs RGB and NIR channels under shared geometry, ensuring spectral and spatial coherence across bands. To support research in this direction, we released a controlled multispectral greenhouse dataset and an accompanying few-shot benchmarking package that standardizes sparse-view evaluation protocols.

Overall, LGDWT-GS demonstrates that incorporating frequency-domain priors and multispectral supervision is an effective strategy for constructing efficient, detail-preserving 3D scene representations under data-limited conditions. Future work will focus on extending this framework with frequency-guided densification and pruning strategies to adaptively refine Gaussian distributions based on spectral frequency cues and to generalize the approach to broader multispectral datasets. Additionally, we aim to extend this framework to “in-the-wild” or less controlled agricultural environments, such as field-grown crops, to enable robust real-world 3D reconstruction under challenging outdoor conditions~\cite{zhang2025wheat3dgs}.

\section*{Acknowledgment}
This material is based upon work supported by the Texas A\&M University System National Laboratories Research Seed Funding Program. Any opinions, findings, and conclusions or recommendations expressed in this material are those of the author(s) and do not necessarily reflect the views of the Los Alamos National Laboratory or The Texas A\&M University System
The authors also acknowledge the use of ChatGPT (OpenAI) for assistance with language polishing and grammar refinement.

\bibliographystyle{IEEEtran}
\bibliography{egbib}


\begin{IEEEbiography}[{\includegraphics[width=1in,height=1.25in,clip,keepaspectratio]{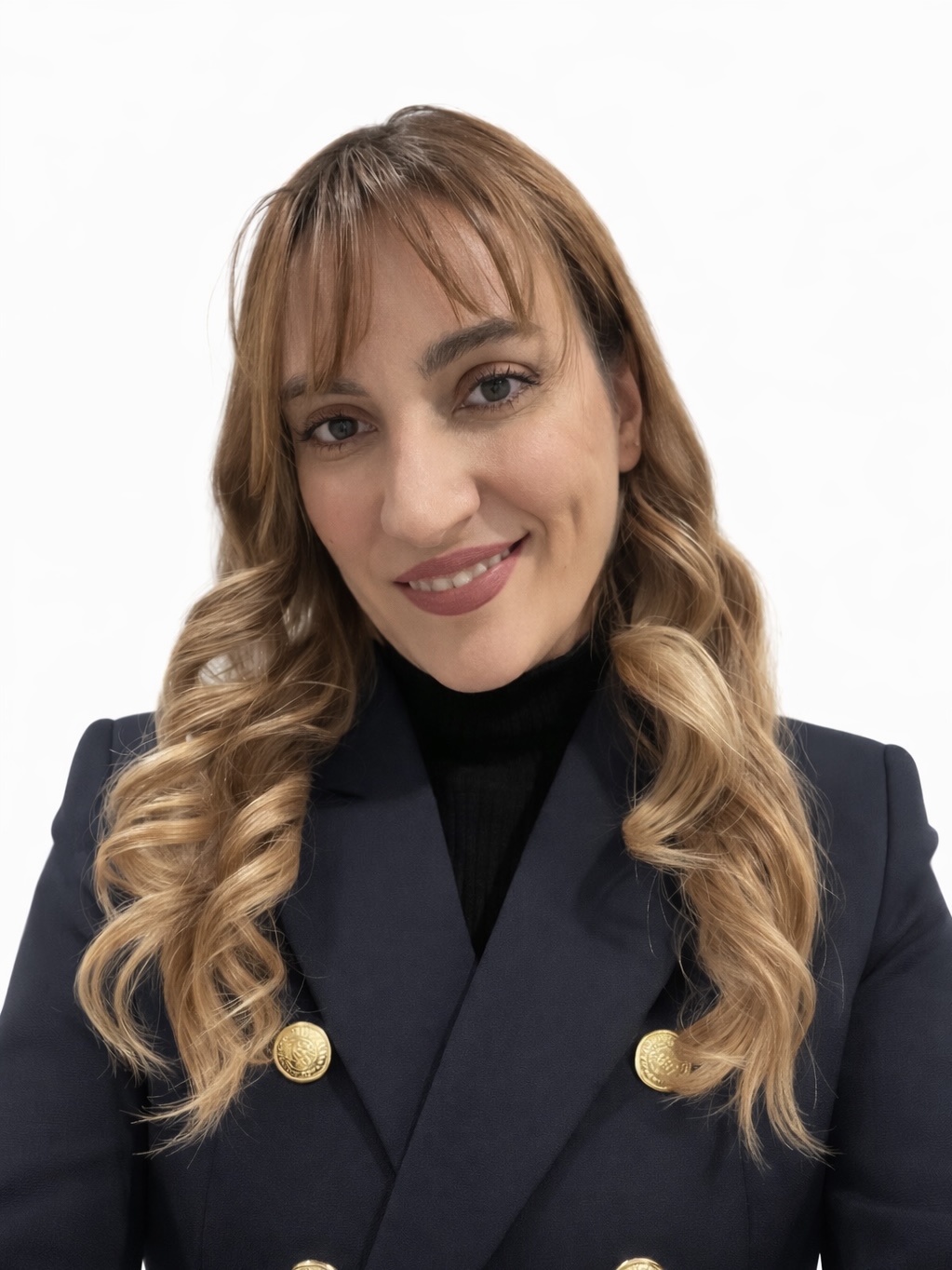}}]{Shima Salehi}
received the B.Sc. degree in electrical engineering from Isfahan University of Technology in 2020 and the M.Sc. degree in computer science from Amirkabir University of Technology in 2023. She is currently pursuing the Ph.D. degree in computer engineering at Texas A\&M University. Her research focuses on sparse-view 3D reconstruction and neural scene representations, emphasizing geometric consistency and structural fidelity in real-world environments.
\end{IEEEbiography}
\begin{IEEEbiography}[{\includegraphics[width=1in,height=1.25in,clip,keepaspectratio]{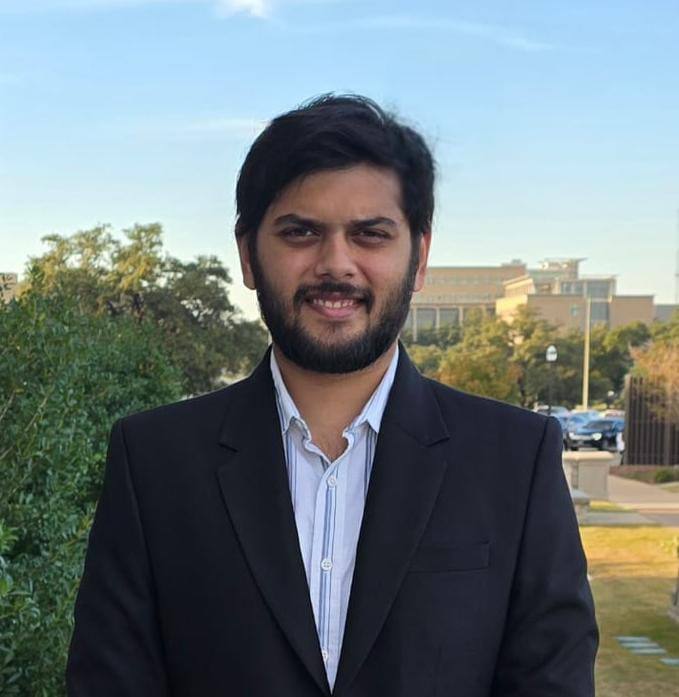}}]{Atharva Agashe}
received the B.E. degree in electronics and telecommunications from the Pune Institute of Computer Technology, Pune, India. He received the M.S. degree in computer engineering from Texas A\&M University. His research interests include acoustics, signal processing, computer vision, self-supervised learning, and 3D reconstruction.
\end{IEEEbiography}
\begin{IEEEbiography}[{\includegraphics[width=1in,height=1.25in,clip,keepaspectratio]{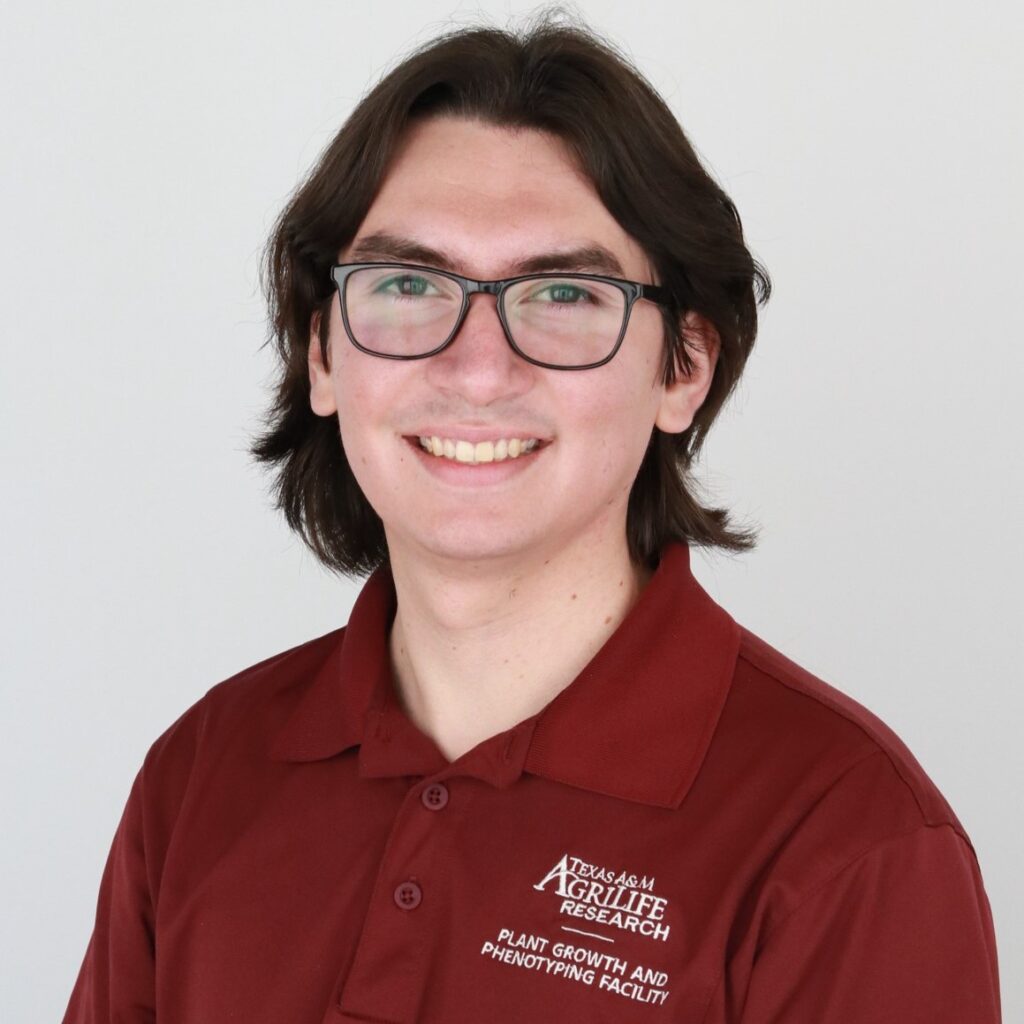}}]{Andrew McFarland}
received the B.S. degree in plant and environmental soil science from Texas A\&M University in 2022. He is currently a Research Assistant with the Texas A\&M AgriLife Research Plant Growth and Phenotyping Facility and is pursuing the M.S. degree in horticultural sciences at Texas A\&M University. His research focuses on grapevine phenotyping under salinity stress.
\end{IEEEbiography}
\begin{IEEEbiography}[{\includegraphics[width=1in,height=1.25in,clip,keepaspectratio]{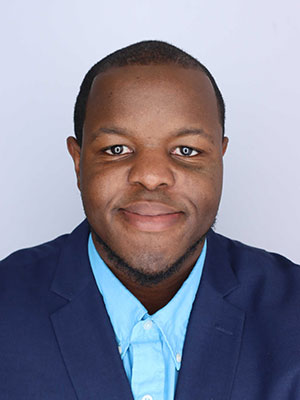}}]{Joshua Peeples}
received the Ph.D. degree in electrical and computer engineering from the University of Florida in 2022. He is an Assistant Professor in the Department of Electrical and Computer Engineering at Texas A\&M University. His primary research interests include machine learning, computer vision, and image processing with a focus on image texture analysis.
\end{IEEEbiography}

\EOD
\end{document}